\documentclass[a4paper]{paper}
\usepackage[left=1in,right=1in]{geometry}
\usepackage[
backend=biber,
style=authoryear,
sorting=nyt,
useprefix
]{biblatex}
\DeclareSortingNamekeyTemplate{
  \keypart{
    \namepart{family}
  }
  \keypart{
    \namepart{prefix}
  }
  \keypart{
    \namepart{given}
  }
  \keypart{
    \namepart{suffix}
  }
}
\usepackage[table,dvipsnames]{xcolor}
\usepackage[colorlinks,citecolor=ForestGreen]{hyperref}
\usepackage{parskip}
\usepackage{booktabs}
\usepackage{graphicx}
\usepackage{rotating}
\usepackage{subfig}
\usepackage{pgfplotstable}
\pgfplotsset{compat=1.15}
\usepackage{etoolbox}
\usepackage{comment}
\usepackage{framed}
\usepackage[export]{adjustbox}
\usepackage{authblk}

\pgfplotstableset{
    color cells/.style={
       assign column name/.style={
         /pgfplots/table/column name={\small ##1}
       },
       col sep=comma,
       string type,
       postproc cell content/.code={%
             \pgfkeysalso{@cell content=\rule{0cm}{2.4ex}%
                  \ifstrequal{##1}{}{}{%
                      \pgfmathtruncatemacro\number{##1}%
                      \pgfmathtruncatemacro\scalecolor{##1*6+5}%
                      \edef\temp{\noexpand\cellcolor{blue!\scalecolor}}\temp%
                      \ifnum\number>6\textcolor{white}{\bfseries##1}%
                      \else##1\fi
                      }%
             }},
        columns/x/.style={
            column name={},
            postproc cell content/.code={}
        }
    }
}

\newcommand{\beginsupplement}{%
        \setcounter{table}{0}
        \renewcommand{\thetable}{S\arabic{table}}%
        \setcounter{figure}{0}
        \renewcommand{\thefigure}{S\arabic{figure}}%
     }

\title{Negation detection in Dutch clinical texts: an evaluation of rule-based and machine learning methods}
\author[1,8,*]{Bram van Es}
\author[2]{Leon C. Reteig}
\author[3]{Sander C. Tan}
\author[4]{Marijn Schraagen}
\author[5]{Myrthe M. Hemker}
\author[6]{Sebastiaan R.S. Arends}
\author[7]{Miguel A.R. Rios}
\author[1]{Saskia Haitjema}
\affil[1]{Central Diagnostic Laboratory, University Medical Center Utrecht, Utrecht University, the Netherlands}
\affil[2]{Center for Translational Immunology, University Medical Center Utrecht}
\affil[3]{Department for Research \& Data Technology, University Medical Center Utrecht}
\affil[4]{Institute for Information and Computing Sciences, Utrecht University}
\affil[5]{Utrecht Institute of Linguistics OTS \& Department of Languages, Literature and Communication, Utrecht University}
\affil[6]{Department of Medical Informatics, University of Amsterdam}
\affil[7]{Centre for Translation Studies, University of Vienna}
\affil[8]{MedxAI}
\affil[*]{Corresponding author: b.vanes-3@umcutrecht.nl}

\begin{document}
\maketitle
\begin{abstract}

As structured data are often insufficient, labels need to be extracted from free text in electronic health records when developing models for clinical information retrieval and decision support systems. One of the most important contextual properties in clinical text is negation, which indicates the absence of findings. We aimed to improve large scale extraction of labels by comparing three methods for negation detection in Dutch clinical notes. We used the Erasmus Medical Center Dutch Clinical Corpus to compare a rule-based method based on ContextD, a biLSTM model using MedCAT and (finetuned) RoBERTa-based models. We found that both the biLSTM and RoBERTa models consistently outperform the rule-based model in terms of F1 score, precision and recall. In addition, we systematically categorized the classification errors for each model, which can be used to further improve model performance in particular applications.
Combining the three models naively was not beneficial in terms of performance. We conclude that the biLSTM and RoBERTa-based models in particular are highly accurate accurate in detecting clinical negations, but that ultimately all three approaches can be viable depending on the use case at hand.





\end{abstract}

\section{Introduction}
\label{sec:introduction}
The increasing availability of clinical care data, affordable computing power, and suitable legislation provide the opportunity for (semi-)automated decision support systems in clinical practice. An important step in the development of such a decision support system is the accurate extraction of relevant labels to train the underlying models. These labels are rarely directly available as structured data in electronic health records (EHRs)---and even if they are, they often lack the  precision and reliability \parencite{Stausberg2008} for a clinical decision support system. Therefore, extraction of labels from free text in the EHR---which contains the richest information and the appropriate amount of nuance---is needed. 

To this end, we need to consider the context in which medical terms are mentioned. One of the most important contextual properties in clinical text is \textit{negation}, which indicates the \textit{absence} of findings such as pathologies, diagnoses, and symptoms. As they make up an important part of medical reasoning, negations occur frequently: one study estimated that more than half of all medical terms in certain clinical text are negated \parencite{Chapman_evaluation}. Accurate negation detection is critical when labels and features from free text in the EHR are extracted for use in clinical prediction models. But improving information retrieval through negation detection has many other use cases in healthcare, including administrative coding of diagnoses and procedures, characterizing medication-related adverse effects, and selection of patients for inclusion in research cohorts.

Negation detection is not a trivial task, due to the large variety of ways negations are expressed in natural language\footnote{For example, the rule-based system used in this work contains nearly 400 different patterns of expressing negation.}. It can either be performed with a rule-based approach or through machine learning. In this paper, we evaluate the performance of one rule-based method (based on ContextD \parencite{Afzal2014}) and two machine learning methods (a bidirectional long-short term memory model implemented in MedCat \parencite{Kraljevic2020}, and a RoBERTa-based \parencite{Liu2019} Dutch language model) for detection of negations in Dutch clinical text.

In their simplest form, traditional rule-based methods consist of a list of regular expressions of negation triggers (e.g. ``no evidence for", ``was ruled out"). When a negation trigger occurs just before or after a medical term in a sentence, the medical term is considered negated. Examples include NegEx \parencite{Chapman_algorithm}, NegFinder \parencite{Mutalik2001}, NegMiner \parencite{Elazhary2017} and ConText \parencite{Harkema2009,Shi2018}. Some approaches also incorporate the grammatical relationships between the negation and medical terms. For example, incorporating part-of-speech tagging to determine the noun
phrases that a negation term could apply to (see e.g. NegExpander \parencite{Aronow1999}) 
, or using dependency parsing to uncover relations between words (see e.g. NegBio \parencite{Peng2018}, negation-detection \parencite{Gkotsis2016}, DepNeg \parencite{Sohn2012} and DEEPEN \parencite{Mehrabi2015}). Moreover, distinguishing between the different types of negations (syntactic, morphological, sentential, double negation) as well as adding word distance has been proven helpful (see e.g. NegAIT \parencite{Mukherjee2017} and \textcite{Slater2021}). While usually tailored for English, some of these methods have been adapted for use in other languages, including French \parencite{Deleger2012}, German \parencite{Cotik2016} and Spanish \parencite{Cotik2016_2,Costumero2014}, as well as Dutch \parencite{Afzal2014}.

The main \textit{advantages} of rule-based negation detection methods are that they are transparant and easily adaptable and do not require any scarce labeled medical training data. Rule-based methods can be surprisingly effective. \textcite{Goryachev2006} demonstrate that the relatively simple NegEx can be more accurate than machine learning-based methods, such as a Support Vector Machine (SVM) trained on the part-of-speech-tags surrounding the term of interest.

The main \textit{disadvantage} of rule-based methods is that they are by definition unable to detect negations that are not explicitly captured in a rule. Depending on the use case, this can severely hamper their performance. This is where machine learning methods come into play, as they may outperform rule-based methods by picking up rules implicitly from annotated data.

One such machine learning method is the bidirectional long-short term memory model (biLSTM), a neural network architecture that is particularly suited for classification of sequences such as natural language sentences. This model processes all words in a sentence sequentially, but in contrast to traditional neural network methods, a biLSTM takes the output of previous words into account to model relations between words in the sentence. For a biLSTM model the processing is bidirectional, meaning that sentences are processed in the (natural) forward direction, as well as the reverse direction.

Based on a conventional biLSTM (see e.g. \textcite{Graves2005}), \textcite{Sun2019} developed a hybrid biLSTM Graph Convolutional Network (biLSTM-GCN) method. The biLSTM can also be combined with a conditional random field \parencite{Huang2015}. Other machine learning approaches include an SVM that has access to positional information of the tokens in each sentence \textcite{Cruz2012}, and the use of conditional random fields in the NegScope system by \textcite{Agarwal2010}.

A more recent machine learning model is RoBERTa \parencite{Liu2019}, a bidirectional neural network architecture that is pre-trained on extremely large corpora using a self-supervised learning task, specifically to fill in masked tokens. This masking does not require external knowledge as the selection of tokens to be masked can be performed automatically. RoBERTa is part of a family of models, which primarily vary in learning task, that are based on the transformers architecture \parencite{Vashwani2017}. Once pre-trained, a transformer model can be finetuned with supervised learning tasks for e.g. negation detection or named entity recognition. \textcite{Lin2020} show that a zero-shot language model such as BERT performs well on the negation task and does not require domain adaptation methods. \textcite{Khandelwal2020} developed NegBERT, a BERT model finetuned on open negation corpora such as BioScope \parencite{Vincze2008}.



The goal of the current paper is to compare the performance of rule-based and machine learning methods for Dutch clinical data. We conduct an error analysis of the types of errors the individual models make, and also explore whether combining the methods through ensembling offers additional benefits. Python implementations of all the evaluated models are available on GitHub\footnote{\url{https://github.com/umcu/negation-detection/releases/tag/v1.0.0}}.

\section{Data}
We used the Erasmus Medical Center Dutch clinical corpus (DCC) collected by \textcite{Afzal2014} (published together with ContextD) that contains 7490 Dutch anonymized medical records annotated with medical terms and contextual properties. All text strings that exactly matched an entry in a subset of the Dutch Unified Medical Language System (UMLS, \parencite{bodenreider2004unified}) were considered a medical term. These medical terms were subsequently annotated for three contextual properties: temporality, experiencer, and negation. In this paper we focus on the binary context-property negation. The label \textit{negated} was given when evidence in the text was found that indicated that a specific event or condition did not take place or exist, otherwise the label \textit{not negated} was assigned.

\begin{figure}[ht]
\centering
\includegraphics[width=\textwidth]{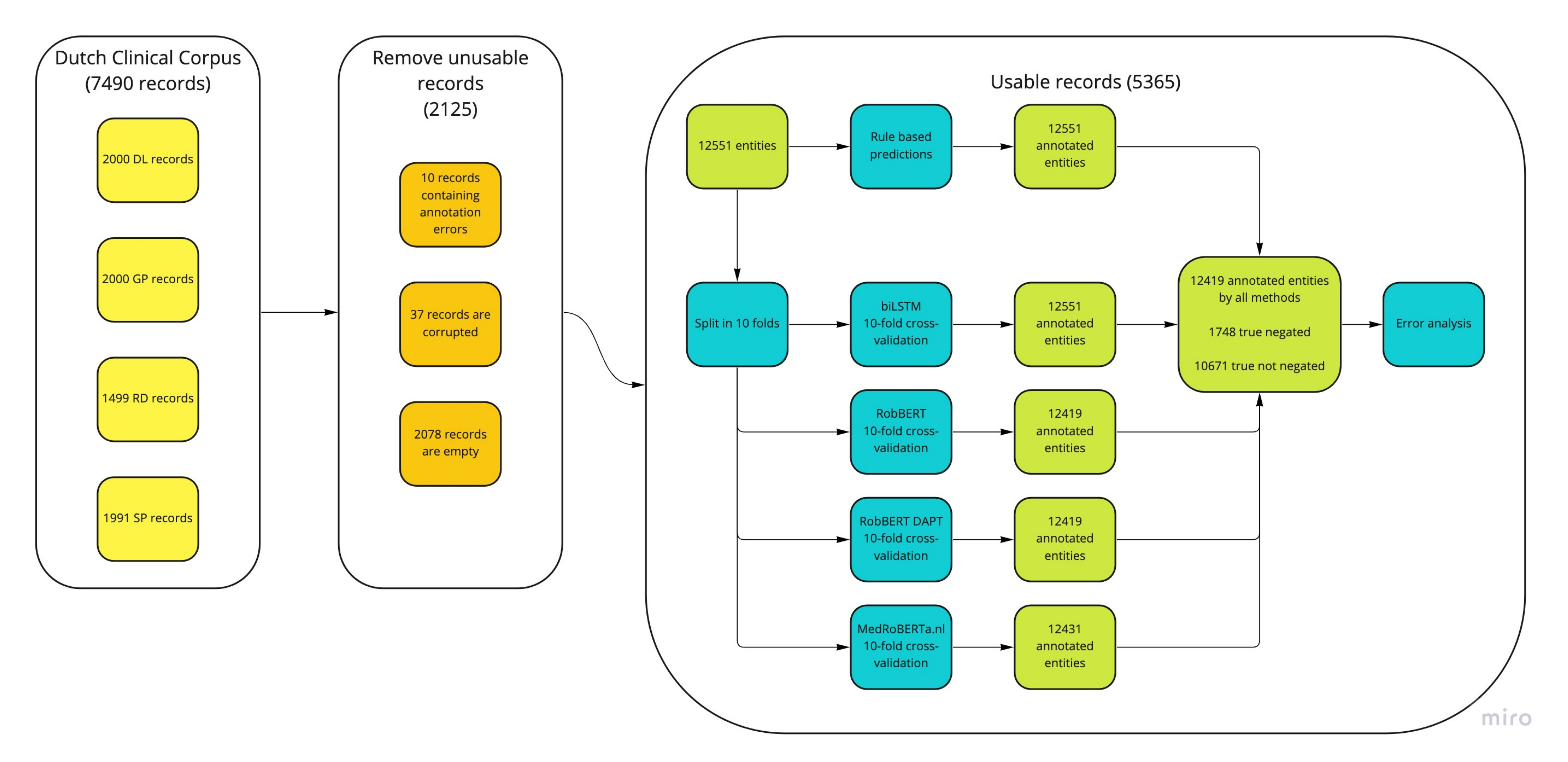}
\caption{Data flow diagram}
\label{fig:data_flow_diagram}
\end{figure}

As illustrated in Figure~\ref{fig:data_flow_diagram}, we excluded 2125 records in total from further analysis, primarily because no annotation was present (this was the case for 2078 records), otherwise the file containing the source text or its annotation was corrupted (37 records), or the annotation did not correspond to a single medical term (10 records, e.g. only a single letter was annotated, or a whole span of text containing multiple medical terms). This left 5365 usable records for analysis, containing a total of 12551 annotated medical terms. A small number of medical terms were not processed by the RoBERTa-based models, because of the imposed maximum record length of 512 tokens in our implementation. We excluded these medical terms from analysis with other methods as well, resulting in a final set of 12419 annotated medical terms.

The corpus consists of four types of clinical records, which differ in structure and intent (for details, see \textcite{Afzal2014}). Basic statistics are presented in Table~\ref{tab:basic_properties_records} and a representative example of each record type, including various forms of negation, is provided in Supplementary Material \ref{sec:text-examples}.

\begin{table}[ht]
\begin{tabular}{lllll}
\toprule
Letter category & \# sentences & \# words & \# unique words & word length \\
\midrule
General Practitioner & 2.1 (1, 2) & 17.8 (8, 23) & 16.8 (8, 22) & 5.9 (5, 6.5)\\
entries              & 2034       & 33840        & 11080        & \\    
\midrule
Specialist letters   & 3.8 (2, 4) & 30.0 (9, 34) & 24.9 (8, 30) &  7 (5.8, 7.7)\\
                     & 2737       & 29674        & 10207        & \\
\midrule
Radiology reports    & 3.6 (2, 4) & 19.1 (7, 26) & 16.8 (6, 23) &  7.2 (6.1, 8)\\
                     & 3939       & 28614        & 6371         & \\
\midrule
Discharge letters    & 4.1 (2, 5) & 33.8 (14, 45)& 27.8 (13, 38)& 6.1 (5.5, 6.6) \\
                     & 3057       & 33458        & 6351         & \\
\bottomrule
\end{tabular}
\caption{\textbf{Basic textual statistics of the selected DCC records}, showing the mean value per record and the boundaries of the second and third quartile (top) and the total count in the dataset (bottom).}
\label{tab:basic_properties_records}
\end{table}

\section{Methodology}




We employed three distinct methods to identify negations: a rule-based approach called ContextD, a biLSTM from MedCAT and a fine-tuned Dutch RoBERTa model. These methods are cross-validated using the same ten folds.

\subsection{Rule-based approach}
The rule-based ConTextD algorithm \parencite{Afzal2014} is a Dutch adaptation of the original ConText algorithm \parencite{Harkema2009}.

The backbone of the ConText algorithm is a list of (regular expressions of) negation terms (``negation triggers"). A given medical term is considered to be negated when it falls within the scope of a negation trigger. The default scope in ConText is the remainder of the sentence after/before the trigger. Each negation trigger has either a forward (e.g., ``no evidence of ...") or backward scope  (e.g., ``... was ruled out").

ConText has two more types of triggers aside from negation triggers: pseudo-triggers and termination triggers. Pseudo-triggers are phrases that contain a negation trigger but should not be interpreted as such (e.g., ``not only" is a pseudo-trigger for the negation trigger ``not"). Pseudo-triggers take precedence over negation triggers: when a pseudo-trigger occurs in a sentence, the negation trigger it encompasses is not acted upon. Termination triggers serve to restrict the scope of a negation trigger. For example, words like ``but" in the sentence ``No signs of infection, but pneumonia persists", signal that the negation does not apply to the entire sentence. Using ``but" as termination trigger prevents the algorithm to consider ``pneumonia" to be negated while ``infection'' is still considered negated.

We used the Dutch translation of the original triggers from ConText, as produced by the ContextD authors\footnote{\href{https://biosemantics.erasmusmc.nl/ContextD/translation/NL.txt}{Erasmus Medical Center website}}. These triggers were used in conjunction with MedSpaCy\footnote{\href{https://github.com/medspacy/medspacy/releases/tag/0.1.0.2}{medspacy, version 0.1.0.2}} \parencite{Eyre_medspaCy}, a Python implementation of the ConText algorithm. Because ConText defines the scope of a negation trigger in number of words or the boundary of the sentence (the default), raw text also needs to be tokenized and split into separate sentences. We used the default tokenizer and the dependency-parser-based sentence splitter of the \verb|nl_core_news_sm-2.3.0| model in spaCy\footnote{\href{https://v2.spacy.io}{spaCy, version 2.3.5}}, a generic Python library for NLP.

\subsection{MedCAT's biLSTM}
The open-source multi-domain clinical natural language processing tool Medical Concept Annotation Toolkit (MedCAT) incorporates Named Entity Recognition (NER) and Entity Linking (EL), to extract information from clinical data \parencite{Kraljevic2020}.
MedCAT contains a component named MetaCAT for dealing with context properties after a concept is identified in a clinical text. MetaCAT makes use of bidirectional Long-Short Term Memory networks (biLSTMs) \parencite{Graves2005}. This sequence encoding can be combined directly with a classifier such as a fully connected network (FCN) or indirectly via a Conditional Random Field or Graph Convolutional Network to facilitate interaction between entities. In this work we use an FCN for the final classification. By MetaCAT default, the biLSTM observes 15 tokens to the left and 10 tokens to the right of the annotated concept.

MetaCAT replaces target terms, for example \textit{diabetes mellitus}, with an abstract placeholder, such as \textit{[disease]}. This allows the model to learn a shared representation between concepts, which increases the generalization behavior of the model for negation detection. For example, the model now only needs to learn that \textit{no signs of [disease]} is a negation, instead of learning \textit{no signs of diabetes mellitus}, \textit{no signs of bronchitis}, ... for each disease separately.

MetaCAT uses a Byte Pair Encoding (BPE) tokenizer which operates on a subword level \parencite{Gage1994}. Earlier research \parencite{Mascio2020} showed that the subword tokenizer outperforms traditional tokenizers that operate on word level. When training MetaCAT's biLSTM, embeddings are required. For this reason, first a set of Continuous Bag of Words (CBOW) embeddings were created using a set of Dutch Wikipedia articles in the medical domain. Secondly, the biLSTM model is trained on the annotated DCC data using the default MetaCAT settings.

While MedCAT provides a full pipeline including NER and entity linking, in the current project we only use the biLSTM negation classifier component, using the entities marked in the DCC as input. This allows for easy comparison with the other two methods that do not share the same pipeline architecture.

\subsection{RoBERTa}
RoBERTa \parencite{Liu2019} is part of a family of language models built on top of the transformers architecture that allows for learning general contextual representations using self-supervised training. The variation in models such as BERT \parencite{Devlin2019}, XLM \parencite{Lample2019} and T5 \parencite{Raffel2020} primarily comes from differences in training objectives.

A benefit of RoBERTa, BERT and similar transformer-based methods is that the context of a term can play an important role. With sequential models such as LSTMs long range inter-word dependencies are hardly taken into account for the simple reason that direct neighbors have more influence on eachother\footnote{Recurrent neural networks suffer from the exploding and vanishing gradient problem, with LSTMs and GRUs this is resolved basically through gates between sequence elements and weight clipping. These interventions allow for larger sequences but cannot prevent that there is a monotonic decline in influence away from each token.}. Bidirectional LSTM's alleviate the uni-directionality but still suffer from the inability to incorporate large contexts.

A possible downside of using a pre-trained model is that the standard version of RoBERTa is trained on general text, which may not perform well in the medical domain.

The RobBERT language model \parencite{Delobelle2020} is a RoBERTa architecture pretrained on the Dutch SoNaR corpus (see \textcite{Oostdijk2013}) using Masked Language Modeling, a self-supervised learning process where the model learns to fill in masked tokens. SoNaR consisted primarily of texts obtained from publicly available media. 
MedRoBERTa.nl \parencite{Verkijk2021} is a RoBERTa model that was trained from scratch on 11.8GB of EHRs from the Dutch AUMC tertiary care center. 

Following \textcite{Lin2020}, we compared RobBERT with a domain-adapted (DAPT) RobBERT. Here the domain-adaptation is performed by continued pre-training of the model on a domain-specific corpus consisting of Dutch Medical Wikipedia, NHG directives and standards\footnote{\href{https://www.nhg.org/}{https://www.nhg.org/}} (medical guidelines for general practitioners), Richtlijnendatabase \footnote{\href{https://demedischspecialist.nl/}{https://demedischspecialist.nl/}} (medical guidelines for hospital specialties) and Huisarts en Wetenschap\footnote{\href{https://www.henw.org/}{https://www.henw.org/}} (monthly magazine of the Dutch general practitioners' association, issues 1957-2019). \\ \\
We finetuned all RoBERTa models on the annotated DCC data for 3 epochs with a batchsize of 32 and 64 samples. We performed the finetuning in two ways; all layers are updated or only the final classification layer is trained for the negation detection task. The latter is also called self-supervised plus simple fit (or SSS).

\begin{figure}[ht]
    \centering
    \includegraphics[width=0.7\textwidth]{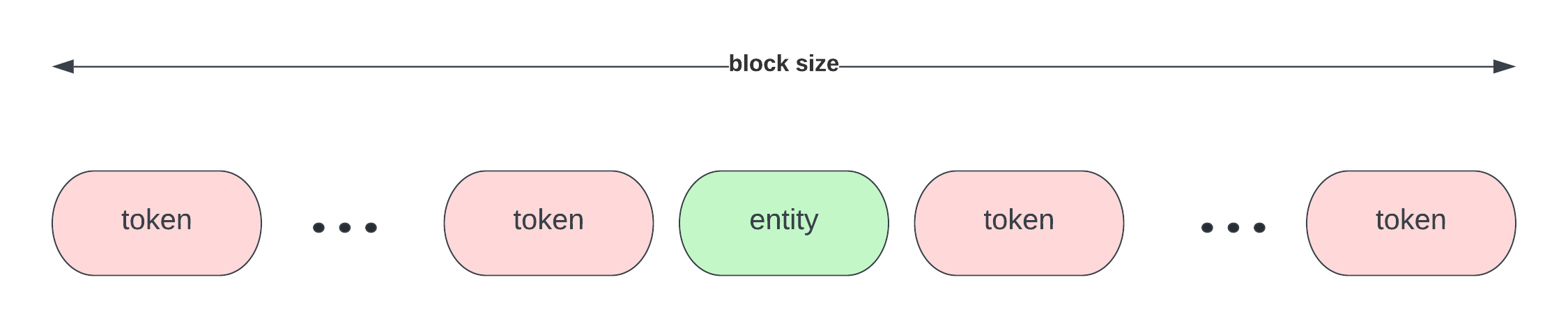}
    \caption{Variable size token window}
    \label{fig:token_window}
\end{figure}

Finally we considered the effect of decreasing the maximum number of tokens, i.e. the maximum sequence length. Decreasing the maximum sequence length will linearly decrease both the space and time complexity of the model training. We varied the number of tokens by taking a token window around each entity (i.e. an annotated medical term in the DCC) (Figure~\ref{fig:token_window}). For the training this means that entities may occur in multiple token windows per record. For the validation we only use the entities in the center of each token window.

\subsection{Ensemble classification}
To investigate whether the three models could complement each other, we created an ensemble classifier. The classifier used a majority voting ensemble, which assigns the label that is predicted by at least two of the three classifiers.

\subsection{Model evaluation}
To gauge the performance of the models we applied 10-fold cross-validation. We inspected the errors in the test folds and assigned error categories facilitating inter-model comparison (see section \ref{sec:error-analysis}). This setup also allows for easy evaluation of the proposed ensemble classifier post-hoc based on the predictions of the individual methods.
 
We evaluated each model's recall, precision, and F1-score for the \textit{negated} class. Recall (true positive rate or sensitivity) is defined as the proportion of true positives out of the sum of true positives and false negatives, and thus indicates what proportion of all negations a model correctly identified. Precision (positive predictive value) is defined as the proportion of true positives out of all positives, and thus indicates how often the model was correct when it predicted that a term was negated. The F1-score is the harmonic mean of precision and recall.

\subsection{Error analysis}
\label{sec:error-analysis}
To better understand the types of misclassifications that the models made, we reviewed all the false positives and false negatives separately for each of the three models, similar to what \cite{Afzal2014} did to evaluate ContextD. A false negative occurred when the model predicted ``not negated" when a negation was present; a false positive occurred when the model predicted ``negated" when no negation was present.

Error analysis allows to determine the expected gain of further steps aimed at preventing errors, which can be either (1) to apply preprocessing to remove artifacts that negatively influence model performance, (2) to remove inconsistencies in data annotation and/or use synthetic data to train the models, or (3) to combine the different models in a complementary fashion.

We first categorized all errors made by a single model. We then attempted to re-use these categories as much as possible for the error analysis of the remaining two models, only creating new categories when there was a clear need. Each model's errors were reviewed by a single invidual (BvE, LCR, and MS). This initial round was followed by two more rounds of discussion among the three reviewers, lumping and splitting the candidate categories, to obtain a final set of categories that was shared across models. This process resulted in 10 different error categories (Table~\ref{tab:errortypes}). To assess the degree of consistency in categorization of the three reviewers, we computed Cohen's kappa coefficient on the subset of errors that were shared by multiple models.

\begingroup
\renewcommand{\arraystretch}{1.3}
\begin{table}[h]
    \centering
    \begin{tabular}{p{0.2\linewidth} | p{0.7\linewidth}}
    \textbf{Category} & \textbf{Definition and example}\\
    \hline
    uncommon negation & Negation term rarely occurs in the data \newline \textit{... which \underline{argues against} a [diagnosis] of RA}\\
    minus & In clinical notes, a minus directly following a term indicates negation \newline \textit{pale-, [nauseous]\underline{-}, clammy+} \\
    scope & Several medical terms follow/precede a negation, but the negation does not apply to all of them. This often occurs in a list, or simply when the negation occurs much later/earlier in the sentence \newline \textit{\underline{no} abdominal pain, [abrasion on leg]} \\
    punctuation & Punctuation that is likely to hinder sentence splitting and/or correct recognition of the (scope of the) negation term \newline \textit{\underline{no evid. for} [aneurysm]} \\
    negation of different term & The negation applies to another term close to the medical term \newline \textit{\underline{no} further investigation of [weight loss]}\\
    wrong modality & The context is not negation, but hypothetical, historical, or otherwise \newline \textit{indication: to \underline{rule out} [osteopenia]} \\
    speculation & The clinician expressed uncertainty instead of outright negation \newline \textit{\underline{no} [eczema] after all?}\\
    ambiguity & The grammatical structure makes it unclear whether a term is negated or not \newline \textit{...\underline{without} loss of function and [concussion] following...}\\
    other & The type of error does not fit into any of the other categories\\
    annotation error & The original annotator assigned the wrong label\\
    \end{tabular}
    \caption{\textbf{Definitions and examples of the error categories used for error analysis}. Negation terms are underlined; the annotated medical terms are in brackets. Examples translated from the Dutch source text.}
    \label{tab:errortypes}
\end{table}
\endgroup

\section{Results}

We consider the model performance quantitatively by looking at overall performance metrics, and more qualitatively by analyzing and categorizing the errors that each model made. 

\subsection{Overall performance}

From the 12419 medical terms in 5365 medical records, 1748 medical terms were marked as negated by the annotators. Of these, 1687 concepts were identified by at least one of the negation detection models. 

The precision, recall and F1 score for each negation detection method are reported in Table~\ref{tab:method_metric_comparison}. RobBERT achieved the highest scores overall, followed by the ensemble method for a few metrics and record types.

\begin{table}[ht]
\begin{tabular}{llrrr}
\toprule
             Letter category & Prediction method &  Precision &  Recall &    F1 \\
\midrule
           Discharge letters &        Rule-based &      0.893 &   0.921 & 0.906 \\
           Discharge letters &            BiLSTM &      0.957 &   0.931 & 0.944 \\
           Discharge letters &           RobBERT &      0.953 &   \textbf{0.974} & 0.963 \\
           Discharge letters &   Voting ensemble &      \textbf{0.963} &   0.966 & \textbf{0.964} \\
\midrule
General Practitioner entries &        Rule-based &      0.674 &   0.801 & 0.732 \\
General Practitioner entries &            BiLSTM &      0.889 &   0.889 & 0.889 \\
General Practitioner entries &           RobBERT &      \textbf{0.950} &   \textbf{0.912} & \textbf{0.931} \\
General Practitioner entries &   Voting ensemble &      0.930 &   0.886 & 0.908 \\
\midrule
           Radiology reports &        Rule-based &      0.901 &   0.966 & 0.932 \\
           Radiology reports &            BiLSTM &      0.933 &   0.934 & 0.933 \\
           Radiology reports &           RobBERT &      \textbf{0.960} &   0.963 & \textbf{0.961} \\
           Radiology reports &   Voting ensemble &      0.955 &   \textbf{0.965} & 0.960 \\
\midrule
          Specialist letters &        Rule-based &      0.807 &   0.840 & 0.823 \\
          Specialist letters &            BiLSTM &      0.922 &   0.835 & 0.876 \\
          Specialist letters &           RobBERT &      0.934 &   \textbf{0.890} & \textbf{0.911} \\
          Specialist letters &   Voting ensemble &      \textbf{0.937} &   0.862 & 0.898 \\
\midrule
    All letters &   Rule-based &      0.825 &   0.892 & 0.857 \\
    All letters &       BiLSTM &      0.926 &   0.901 & 0.913 \\
    All letters &      RobBERT &      \textbf{0.951} &   \textbf{0.938} & \textbf{0.944} \\
   All letters & Voting ensemble &      0.948 &   0.924 & 0.936 \\
\bottomrule
\end{tabular}
\caption{Classification results across methods and data sources}
\label{tab:method_metric_comparison}
\end{table}

\subsubsection{Rule-based}
While the performance of the rule-based method here was indeed comparable to the original (see the results in Table 5 of \textcite{Afzal2014}), some differences can be identified, that probably arise because we do not use exactly the same rules nor exactly the same dataset. First, there are two different variations of ContextD: the ``baseline" rules, which were simply a translation of the original Context method \parencite{Harkema2009}, and the ``final" rules, which were iteratively adapted using half of the dataset. Our set of rules is most similar to the ``baseline" method, as we chose not to implement any of the modifications in the ``final method" described in the paper. However, our set of rules likely still contains some elements of the ``final method''. Second, while the original ContextD method was evaluated on only half the dataset (as the other half was used for finetuning the rules), we use the full dataset. In our evaluation, performance varied quite strongly over the folds, which indicates that the exact evaluation set that was used influenced the obtained results. Performance varied particularly strongly for the GP entries, which would also explain why the difference in performance with ContextD is largest for this category. 

\subsubsection{Machine Learning (biLSTM, RobBERT)}
The rule-based method is outperformed by both machine learning methods in almost all cases. Its performance varies strongly over the different record types: it performs worst for the least structured records, particularly the GP entries, and best for the most structured records, particularly the radiology reports. The performance gap between the rule-based and machine learning methods shows the same pattern: the less structured the record, the larger the gap.

The biLSTM model consistently outperformed the rule-based approach, except for the radiology reports category, where performance was approximately equal. In turn, RobBERT outperformed the biLSTM model with a difference of 0.02--0.05 in F1 score across record categories.

Additionally, we saw no consistent differences between different RobBERT implementations. The smallest 32-token window resulted in a slightly reduced accuracy, but drastically reduced computational resources (see Tables \ref{tab:robbert_comparison} and \ref{tab:robbert_drilldown}).

Note that for the RobBERT and the biLSTM models we did not apply threshold tuning to optimize for precision or recall---we simply took the default threshold of 0.5 (also note that calibration is required if such a probabilistic measure is applied in clinical practice).

\subsubsection{Model ensemble}
The RobBERT method outperforms the other models as well as the ensemble method when scored on the complete dataset. On the individual categories the ensemble method performs worse or similar to the RobBERT method.

\begin{figure}[ht]
    \centering
    \includegraphics[width=0.75\textwidth]{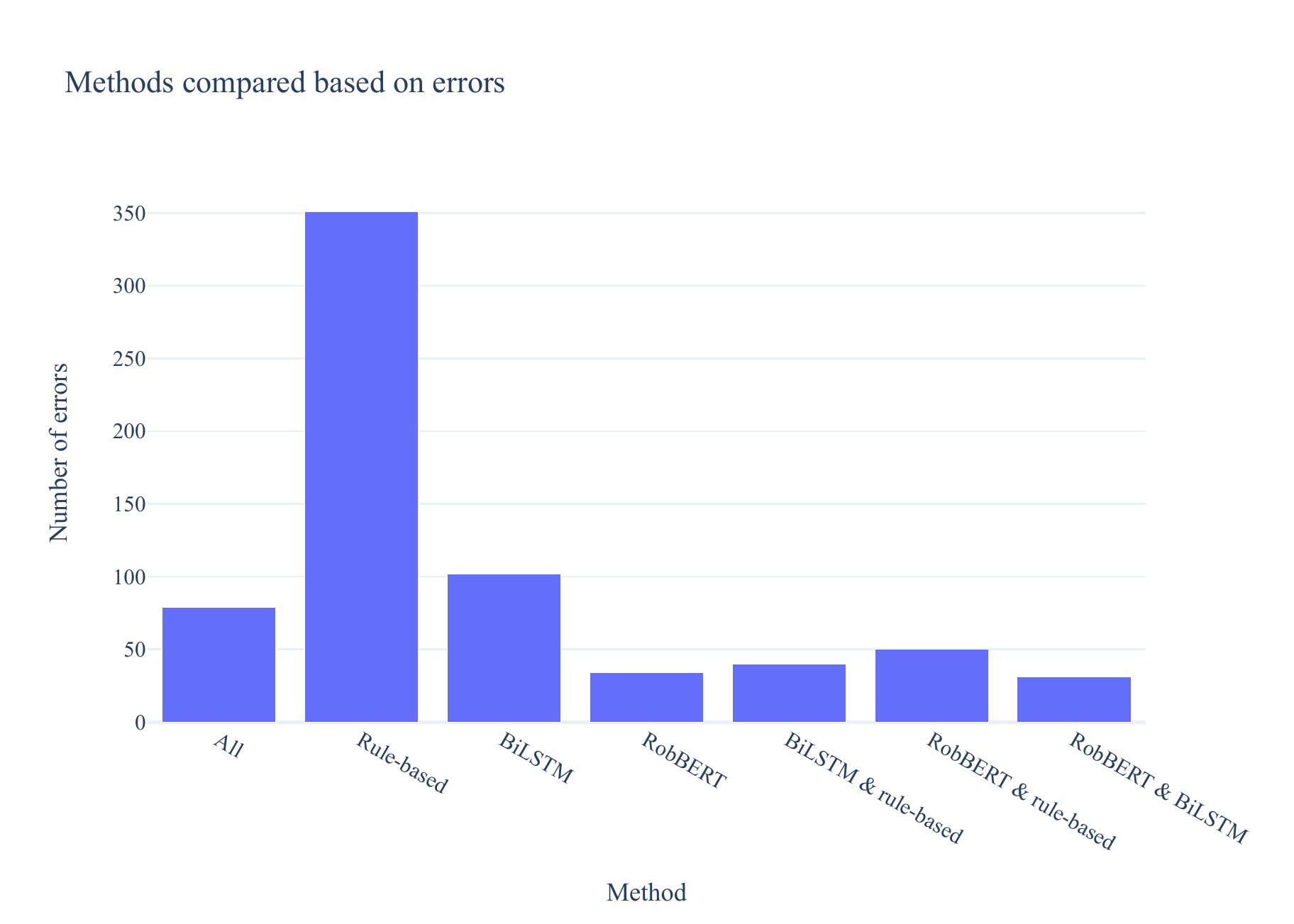}
    \caption{\textbf{Number of misclassifications for different model combinations}. The y-axis shows the number of errors in all possible intersections of the error sets made by the different models. That is, ``All" is the number of entities that are misclassified by \textit{all three} models; ``Rule-based" is the number of entities that are misclassified by \textit{only} the rule-based model; ``RobBERT \& BiLSTM" is the number of entities misclassified by \textit{both} the RobBERT and biLSTM models, \textit{but not} the rule-based model; etc.}
    \label{fig:model_combinations}
\end{figure}

Figure \ref{fig:model_combinations} shows that the RobBERT method (all bars with ``RobBERT") makes fewer errors than the voting ensemble (all bars with more than two methods). In particular, the number of errors committed by RobBERT alone is smaller than the number of errors that are introduced when adding the BiLSTM and the rule-based method to the ensemble.

\subsection{Error analysis}
We obtained a Cohen's Kappa score of $0.48$ when annotating agreement on the error category. This is considered as moderate agreement \parencite{Mchugh2012}. The categories involved in disagreement are shown in Figure~\ref{fig:error_type_combination}. The \textit{uncommon negation} meta-category is a source of disagreement, showing that some annotators consider a negation uncommon while others choose a semantic error category. Also \textit{other} as a (semantic) catch-all category is responsible for disagreement. From the specific categories the \textit{speculation} label is most often subject of disagreement.

The moderate agreement is not surprising given that the error cases represent challenging annotations. It is important to note that it is possible that multiple categories apply for the same error due to model-specific interpretations, which would have a negative effect on our perceived inter-annotator agreement.

\begin{figure}[ht]
    \centering
    \pgfplotstabletypeset[color cells]{
x,Oth,Ann,Spec,Amb,Sc,Pct,Dif,Mod,Minus
Uncommon,11,6,7,5,6,6,1,6,2
Other,,2,7,3,3,7,4,0,0
Annotation,,,3,5,3,1,2,0,0
Speculation,,,,13,3,1,8,3,0
Ambiguous,,,,,2,1,4,1,0
Scope,,,,,,3,3,0,1
Punctuation,,,,,,,1,0,0
Different term,,,,,,,,0,1
Modality,,,,,,,,,0
}
    \caption{Confusion matrix of inter-annotator disagreement}
    \label{fig:error_type_combination}
\end{figure}

\begin{figure}[ht]
    \centering
    \subfloat[][Errors per category]{\includegraphics[width=0.5\textwidth]{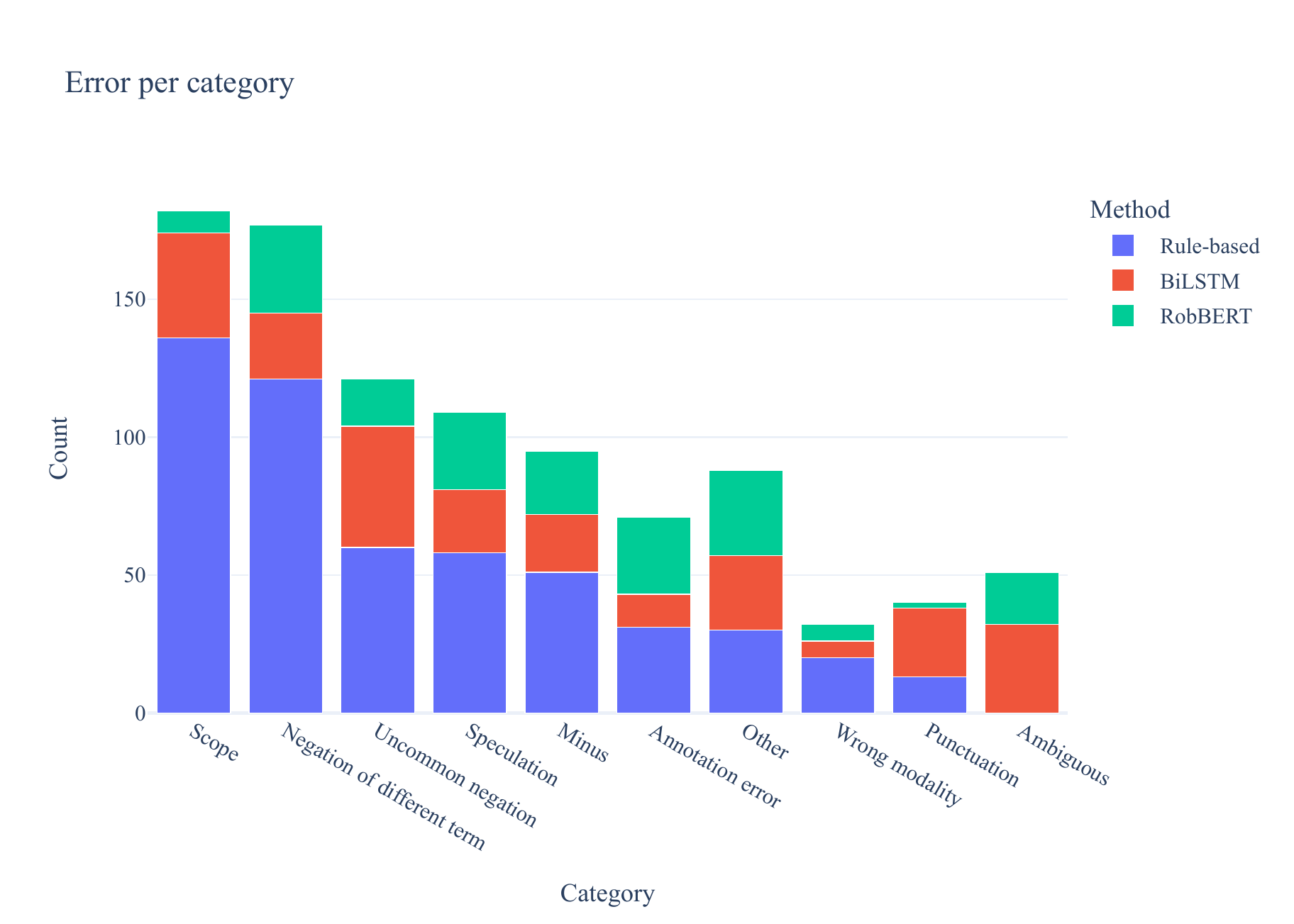}}
    \subfloat[][Errors per method]{\includegraphics[width=0.5\textwidth]{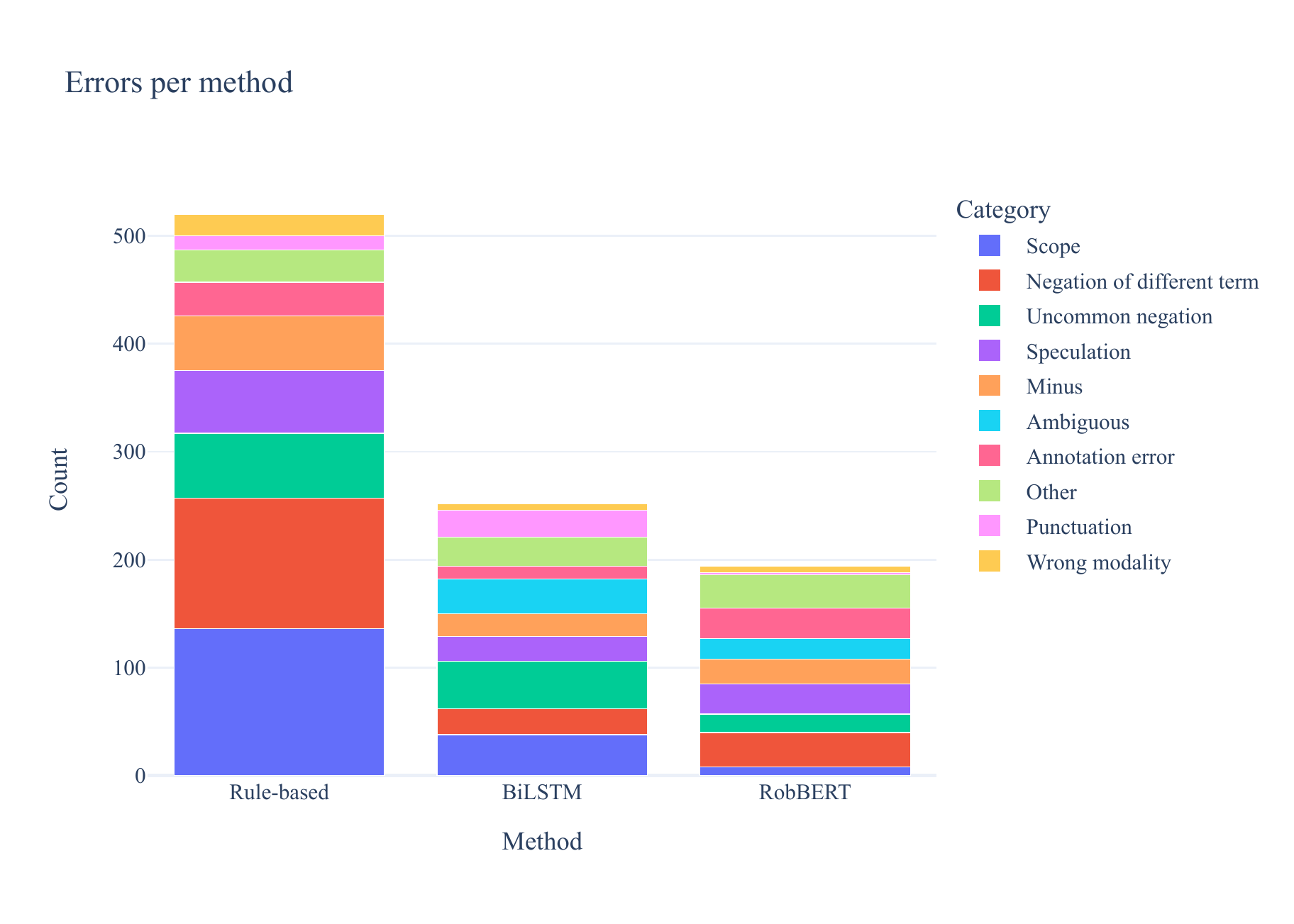}}
    \caption{Error category frequencies}
    \label{fig:model_error_category}
\end{figure}

\begin{table}[!ht]
\centering
\begin{tabular}{lllllll}
\toprule
\multicolumn{1}{l}{\textbf{False positives}} & \multicolumn{2}{l}{Rule-based} & \multicolumn{2}{l}{biLSTM} & \multicolumn{2}{l}{RobBERT} \\
\midrule
Ambiguous                  &          0 &               0\% &     14 &          16\% &      11 &           13\% \\
Annotation error           &         11 &               3\% &      5 &           6\% &      13 &           15\% \\
Minus                      &          0 &               0\% &      8 &           9\% &       0 &            0\% \\
Negation of different term &        119 &              36\% &     24 &          27\% &      32 &           38\% \\
Other                      &          1 &               0\% &     13 &          14\% &      16 &           19\% \\
Punctuation                &          0 &               0\% &      5 &           6\% &       1 &            1\% \\
Scope                      &        136 &              41\% &      6 &           7\% &       0 &            0\% \\
Speculation                &         50 &              15\% &     10 &          11\% &       8 &            9\% \\
Uncommon negation          &          0 &               0\% &      5 &           6\% &       0 &            0\% \\
Wrong modality             &         14 &               4\% &      0 &           0\% &       4 &            5\% \\
Total                      &        331 &              &     90 &          &      85 &           \\
\end{tabular}

\begin{tabular}{lllllll}
\toprule
\multicolumn{1}{l}{\textbf{False negatives}} & \multicolumn{2}{l}{Rule-based} & \multicolumn{2}{l}{biLSTM} & \multicolumn{2}{l}{RobBERT} \\
\midrule
Ambiguous                  &          0 &               0\% &     18 &          11\% &       8 &            7\% \\
Annotation error           &         20 &              11\% &      7 &           4\% &      15 &           14\% \\
Minus                      &         51 &              27\% &     13 &           8\% &      23 &           21\% \\
Negation of different term &          2 &               1\% &      0 &           0\% &       0 &            0\% \\
Other                      &         29 &              15\% &     14 &           9\% &      15 &           14\% \\
Punctuation                &         13 &               7\% &     20 &          12\% &       1 &            1\% \\
Scope                      &          0 &               0\% &     32 &          20\% &       8 &            7\% \\
Speculation                &          8 &               4\% &     13 &           8\% &      20 &           18\% \\
Uncommon negation          &         60 &              32\% &     39 &          24\% &      17 &           16\% \\
Wrong modality             &          6 &               3\% &      6 &           4\% &       2 &            2\% \\
Total                      &        189 &                &    162 &          &     109 &           \\
\bottomrule
\end{tabular}
\caption{Overview of error categories per model}
\label{tab:erroranalysis_all}
\end{table}

We should note that the \textit{speculation} category is somewhat domain specific: a clinician might have observed that a particular diagnostic test yielded no indications for a particular diagnosis; arguably this is not a negation of the diagnosis but merely the explicit absence of a confirmation. These signals may or may not be considered as negations, depending on, for example, whether a test is seen as conclusive or if instead additional testing is required for establishing a diagnosis. This reasoning would require the integration of external knowledge, for example through UMLS. More broadly, the dichotomisation of negation/not-negated is perhaps too coarse given the high prevalence of explicitly speculative qualifications in electronic health records. One clear issue with the dichotomy not-negated/negated is that it biases the annotations and models towards the \textit{non-negated} class, because the \textit{negated} label requires explicit negations whereas \textit{non-negated} is everything else, i.e., negations are more strictly constrained. In some cases it is beneficial if the bias is reversed, for instance to obtain affirmations with low false positivity. A mitigation of the non-negation bias is to introduce a model specifically for affirmations/non-affirmations, or indeed a separate label for speculation (see, e.g. \textcite{Vincze2010}).
\newpage
\subsubsection{Model-independent issues}
The distribution of error categories over methods is shown in Figure~\ref{fig:model_error_category} and Table~\ref{tab:erroranalysis_all}.
The categories \textit{annotation error, speculation, ambiguous} (together around 20-25\% across models) may benefit from more specific annotation guidelines. The other errors can be classified as actual mistakes by the model. From these, preprocessing can potentially improve scope and punctuation-related errors. Modality can be improved using special-purpose classifiers similar to the current negation detection classifier. The remaining problematic errors are negation of a different term, uncommon negation, minus, and other errors (around 50\% of all errors across models).

\paragraph{Annotation- and uncertainty-related errors}

A significant amount of annotation errors was found, both for false positives and false negatives. This is consistent with \textcite{Afzal2014}, who report that about $8\%$ of the false positive negations were due to erroneous annotations (no percentage was reported for false negatives). These errors can be solved by improving the annotation, either through better guidelines or by more strict application and post-hoc checking of these guidelines.

Note that annotation errors will also be present in true negatives/positives, which will remain undetected in case the models make the same mistake as the annotators. If these annotation errors are randomly distributed this is a form of label smoothing, and as such the errors could be useful to reduce overfitting. However, it is likely that these annotation errors are not random and are indicative of inherent ambiguity.

The \textit{speculation} and \textit{ambiguous} categories (together around 20-25\% across models) both stem from uncertainty-related issues, either expressed by the clinician (\textit{speculation}) or in the interpretation of the text (\textit{ambiguous}). These may present problems both during annotation and during model training, as the examples do not fully specify a negation, yet the intended meaning can often be inferred. More specific annotation guidelines could reduce these issues to some extent, by ensuring that the examples of a certain category are consistently annotated. However, the models may still be unable to capture such inferences, even if the examples are consistently annotated.

Typical examples for speculation are: \textit{There is no clear [symptom]} or \textit{The patient is dubious for [symptom]}. The English corpora BioScope \parencite{Vincze2008}, GENIA \parencite{Kim2008} and BioInfer \parencite{Pyysalo2007} include uncertainty as a separate label, which may be beneficial for the current dataset as well.

\paragraph{Remaining errors}

The other categories were more syntactic in nature. The word-level syntactic errors \textit{scope, negation of different term}, and \textit{uncommon negation} occur across methods. Other errors are due to the use of a \textit{minus} to indicate negation or usage of colon and semicolon symbols (\textit{punctuation}). The minus sign is however also used as a hyphen (to connect two words), which complicates handling of this symbol both in preprocessing and during model training.

As an example of possible mitigation measures, the following sentence produced a false negative (i.e., a negated term classified as non-negated) for the target term \textit{redness}:

\begin{framed}
Previously an antibiotics treatment was administered(no redness).
\end{framed}

In this example the negation word \textit{no} is concatenated with an opening parenthesis and the previous word, which poses problems for tokenization. Such errors might be avoided by inserting whitespace during preprocessing.

The following sentence shows a false positive from the biLSTM classifier for the term \textit{earache}:

\begin{framed}
since 1 day pain in the right lower lobe andcoughing, mucus, temp to38.2,\\pulmones no abnormality earache and deaf, oam right?
\end{framed}

This is a scoping error, where the negation on the term ``abnormality" is incorrectly extended to the target term ``earache". This issue would be difficult to correct in preprocessing, as it necessitates some syntactic and semantic analysis to normalize the sentence. This would closely resemble the processing by the rule-based system, therefore this is an example where model ensembling could be beneficial. However, given the multiple other textual issues in this sentence (such as missing whitespace, punctuation, and capitalization) a more robust alternative solution might be to maintain a certain standard of well-formedness in reporting, either through automatic suggestions, reporting guidelines, or both.

\subsubsection{Rule-based}
Almost half of the false positives for the rule-based method fell under the ``scope" category (41\%, see Table~\ref{tab:erroranalysis_all}). The default scope for a negation trigger extends all the way to the start (backward direction) or end (forward direction) of a sentence. For long sentences, or when sentences are not correctly segmented, the negation trigger may then falsely modify many medical terms. This occurred particularly often for short and unspecific negation triggers, such as \textit{no} (Dutch: \textit{niet, geen}). Potential solutions include improving sentence segmentation, restricting the number of medical terms that a single negation trigger can modify, restricting the scope to a fixed number of tokens (the solution used by the ContextD ``final" algorithm for certain record types), or restricting the scope by adding termination triggers (the ContextD ``final" algorithm added punctuation such as colons and semicolons as termination triggers for some record types). We determined that 32 out the 140 false positives were caused by a missing termination trigger; adding just the single trigger \textit{wel} (roughly meaning \textit{but}) would have prevented 18 errors (5\%).

Most other false positives were due to ``negation of a different term". These are perhaps more difficult to fix, but in some cases these could be prevented by adding pseudo-triggers that for instance prevent the trigger ``no" from modifying a medical concept when followed by another term (e.g., ``no relationship with").

For the false negatives, the majority are caused by ``uncommon negations", i.e. negation triggers that were missing from the list of rules. A special case that caused a lot of errors was the minus (hyphen) symbol, which in clinical shorthand is often appended to a term to indicate negation. Other missing negation triggers that occurred relatively often were variations on \textit{negative} (n=18, such as \textit{neg}), \textit{not preceded by} (n=8, e.g., \textit{niet voorafgegaan door}), and \textit{argues against} (n=5, e.g., \textit{pleit tegen}). The obvious way to remedy these error categories is to simply add these negation triggers to the rule list, but this may introduce new problems. For instance, adding ``-" as a negation trigger (as was done in the ContextD ``final" algorithm) would negate any word that occurs before a hyphen (e.g., ``infection-induced disease"). More generally, any change aimed at reducing the amount of false negatives (such as adding negation triggers) or false positives (such as restricting scope) is likely to induce a commensurate increase in false positives or negatives, respectively. The list of rules---and thereby the trade-off between recall and precision---will have to be adapted and optimized for each individual corpus and application. 

\subsubsection{BiLSTM}
As shown in Table~\ref{tab:erroranalysis_all}, for the biLSTM classifier around 5\% of the errors are annotation errors, where the model actually predicted the correct label. For false negatives one of the largest categories is scope errors, which includes examples where a list of entities is negated using a single negation term, as well as examples where many tokens are present between the negation term and the medical term. For false positives, \textit{negation of a different term} is a common error, which is problematic for all three methods. However, compared to the other methods the biLSTM has a more even distribution over error categories. The overall performance and the distribution over categories shows that the biLSTM is more robust against syntactic variation than the rule-based model, but not as generalized as RobBERT.

\subsubsection{RobBERT}

\textit{Negation of different terms, speculation, uncommon negation} and the use of a hyphen (\textit{minus}) to indicate negations are the largest potentially resolvable contributors to the RobBERT error categories, totalling to about $50\%$ of the errors (Table~\ref{tab:erroranalysis_all}). 

RobBERT-base fills in missing interpunction, i.e. it expects interpunction based on the corpus it was trained on and in our clinical case we often find that interpunction is missing. A degenerative example is \textit{``The patient is suffering from palpitations, shortness of breath and udema, this can be an indication of {\color{red}{\textbf{$<$mask$>$}}}"}, RobBERT filled in the mask as {\color{red}{\textbf{:}}} (a semicolon).

A large percentage of the false negatives were due to RobBERT mishandling hyphens. We also observed varying model output based on negation triggers being mixed lower/uppercase. Mishandling the hyphens can potentially be resolved by adapting the tokenizer to include the hyphen as a separate token or by adding white space.

\begin{figure}[ht]
    \centering
    \subfloat[][Varying negation-estimated over single word \textit{Coeliakie}]{\includegraphics[width=0.3\textwidth,valign=t]{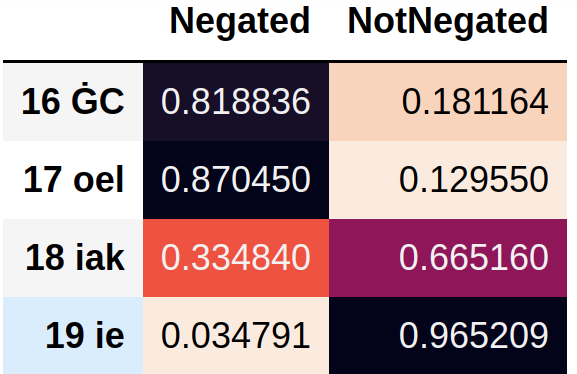}}
    \qquad
    \subfloat[][Varying negation-estimated over single word \textit{Oedeem}]{\includegraphics[width=0.3\textwidth,valign=t]{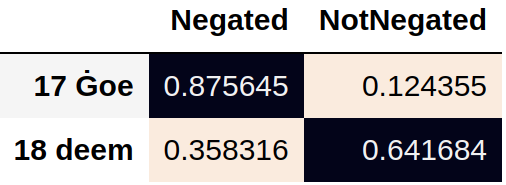}%
    \vphantom{\includegraphics[width=0.3\textwidth,valign=t]{subnegations}}}
    \caption{Intra-word negation variance. The token delimiter character \textbf{\.{G}} is a result of the tokenization}
    \label{fig:varying_estimation}
\end{figure}

We observed that words could have a varying negation estimate over the different tokens that make up the words, illustrated in Figure~\ref{fig:varying_estimation}. This variance is an artefact of using a sub-word tokenizer. This is potentially problematic for words consisting of many tokens, but it also allows for more flexibility because we can decide to (for example) take the maximum probability over the tokens per word. It can also occur that token-specific negations are required, for instance when the negation and the term are concatenated, as in \textit{``De patient is tumorvrij"} (\textit{``The patient is tumorfree"}). The possibility of concatenation is language dependent.

The categories \textit{uncommon negation} and \textit{negation of a different term} can be reduced by expanding the training set with the appropriate samples. 

\section{Discussion}

We compared a rule-based method based on ContextD, a biLSTM model using MedCAT and (finetuned) Dutch RoBERTa-based models on Dutch clinical text and found that both machine learning models consistently outperform the rule-based model in terms of the F1, precision and recall. Combining the three models was not beneficial in terms of performance. The best performing models achieve an F1-score of 0.95. This is a relatively high score for a cross-validated machine learning approach, and is likely near the upper bound of what is achievable for this dataset, considering the noise in labeled data (0.90-0.94 inter-annotator agreement).

\subsection{Applicability}
 
The performance of the assessed methods is well within the acceptable range for use in many information retrieval and data science use-cases in the healthcare domain. Application of these methods can be especially useful for automated tasks where a small number of errors is permitted, such as reducing the number of false positives during cohort selection for clinical trial recruitment. In this task, erroneously excluding a patient is less problematic, and the included patients can be checked manually for eligibility. Other data applications can benefit from this as well, such as text mining for identification of adverse drug reactions, feature extraction for predictive analytics or evaluation of hospital procedures.


However, the model still makes classification errors, which means it is not suitable as a stand-alone method to retrieve automatic annotations for medical decision support systems but can be used directly to improve existing label extraction processes. Application in a decision support system would require some sort of manual interaction with a specialist.

\subsection{Additional aspects for model comparison}
The model comparison (based on precision, recall and F1 score) shows that the RobBERT-based models result in the highest performance. However, additional considerations can play a role in selecting a model, for example computational and human resources. Fine-tuning and subsequently applying a BERT-based model requires significant hardware and domain expertise, which may not be available in clinical practice, or only available outside the medical institution's domain infrastructure, which introduces security and privacy concerns. 

In contrast, the biLSTM and rule-based models can be used on a personal computer, with only a limited performance decrease ($\sim 0.03$ and $\sim 0.08$ respectively) on each evaluation metric. The rule-based method has the advantage that model decisions are inherently explainable, by showing the applied rules to the end user. This may lead to faster adoption of such a system compared to black box neural models. 

The used biLSTM method is part of MedCAT, which also incorporates named entity recognition and linking methods. Compared to the other assessed approaches, this is a more complete end-to-end solution for medical NLP, and is relatively easy to deploy and use, especially in combination with the information retrieval and data processing functionalities of its parent project CogStack \parencite{jackson2018cogstack}. Recently, MedCAT added support for BERT-based models for identification of contextual properties.


\subsection{Limitations and future work}
The study described in this paper has various limitations for which potential improvements can be identified. Regarding language models, the biLSTM network is trained on a relatively small set of word embeddings obtained from Dutch medical Wikipedia articles. This could be complemented or replaced with a larger and more representative dataset, to be more in line with the language models used in the RobBERT experiments. Alternatively, a corpus of actual electronic health records can be used for the best representative dataset, yet this leads to privacy concerns given the high concentration of identifiable protected health information in natural language, even after state of the art pseudonimization.

In the current approach the candidate terms for negation detection in the DCC are generated by performing medical named entity recognition, where each recognized entity is presented to the negation detection models. The context provided to each model, which is assumed be sufficient for determining the presence of negation, is defined to be the sentence around a term as determined by a sentence splitting algorithm. This results in scope-related errors such as incorrect delimitation of the medical term, incorrect sentence splitting, or the negation trigger being in a grammatically different part of the sentence. These issues can be reduced at various stages in the pipeline, either by improving the involved components, or by performing a sanity check on the generated example using part-of-speech-tagging or (dependency) parsing.
Another approach to reduce the number of problematic candidates is to train additional classifiers on meta-properties like temporality \textit{(patient doesn't remember previous occurrences of X)} or experiencer (\textit{X is not common in family of patient}).
Furthermore, the domain-specific structure of the EHR records in the various categories could be leveraged, e.g., to discard non-relevant sections of the health record during processing for specific use cases.

Considering that several error categories are related to the availability of training data, we can to some extent improve the models using synthetic data or a larger set of manually annotated real data.
This can alleviate the lack of balance between the \textit{negation} and \textit{non-negation} classes in the Dutch Clinical Corpus (currently 14\% negations), which are problematic for both the biLSTM and the RobBERT models.
Furthermore, we observe a significant amount of errors due to, or related to, ambiguity. Such errors are expected, not having errors related to ambiguity could indicate an overfitted model.
This idea of error categorisation can also be extended to create a model for estimating the dominant error types in unseen data, i.e. to facilitate model selection and problem-specific model improvements.

In future work it is of interest to train the methods on a broader set of health record corpora, in order to increase the amount of data in general, making it less dependent on DCC specific distributions, and to alleviate the class balance and sparsity issues in particular. 


In this work we compared a small number of methods, and this may have led to a conservative estimate on the performance of the resulting ensemble method. For future work it may be interesting to investigate a bespoke ensemble method where rule-based and machine-learning based methods are combined in a complementary fashion. One technique that is particularly interesting is based on \textit{prompting}, which does not require any finetuning and thus allows pre-trained language models to be leveraged directly.

Unraveling the semantics of clinical language in written electronic health records is a complex task for both algorithms and human annotators, as we experienced during error analysis. However, the three assessed methods show a good performance on predicting negations in the Dutch Clinical Corpus, with the machine learning methods producing the best results. Given the sparse availability of NLP solutions for the Dutch clinical domain, we hope that our findings and provided implementations of the models will facilitate further research and the development of data-driven applications in healthcare.

\section*{Acknowledgements}

We'd like to express our thanks to Jan Kors from the Biosemantics group at ErasmusMC for providing us with the Dutch Clinical Corpus. Also, we thank UMC Utrecht's Digital Research Environment-team for providing high performance computation resources.

\section*{Data availability statement}
The Erasmus Dutch Clinical Corpus dataset can be requested from the Erasmus MC\footnote{\href{https://biosemantics.erasmusmc.nl/index.php/resources/emc-dutch-clinical-corpus}{https://biosemantics.erasmusmc.nl/index.php/resources/emc-dutch-clinical-corpus}}. Code and analysis notebooks are publicly available on GitHub\footnote{\href{https://github.com/umcu/negation-detection}{https://github.com/umcu/negation-detection}} under the MIT license.

\section*{Conflicts of interest}
BvE: None to declare;
LCR: None to declare;
SCT: None to declare;
MS: None to declare;
MMH: None to declare;
SRSA: None to declare;
MARR: None to declare;
SH: None to declare;

\section*{Author contribution}
Conceptualization: BVE; 
Methodology: BVE, LCR, MS, SCT, MMH, SRSA;
Software: BVE, LCR, MS, SCT;
Validation: BVE, LCR, MS, SCT;
Formal analysis: BVE, LCR, MS, SCT, SRSA;
Investigation: BVE, LCR, SCT, MS;
Data Curation: BVE, LCR, SCT, MS, SRSA;
Writing – Original Draft: BVE, LCR, MS, SCT;
Writing – Review \& Editing: BVE, LCR, MS, SCT, MMH, SRSA, MARR, SH;
Visualization: BVE, SRSA;
Supervision: BVE, SH;
Funding acquisition: BVE, SH;

\newpage

\appendix
\section*{Supplementary Material} \label{sec:supplement}
\renewcommand{\thesubsection}{\Alph{subsection}}
\beginsupplement

\subsection{Text examples} \label{sec:text-examples}

We briefly describe the records used for our analysis, in order of least-to-most structured;
 
\textbf{General Practitioner (GP) entries} are clinical notes taken by the GP (a primary care physician) during a patient visit.\\ \\
\textbf{Specialist letters} are longer-form records written by a medical specialist (e.g. a cardiologist), to report back to the GP after their patient has been referred to a hospital. Both of these record types come from the IPCI database \parencite{deRidder_IPCI}, a large collection of medical records from Dutch GPs. The GP entries and (to a lesser extent) the specialist letters are not grammatically well-formed and often contain clinical shorthand. \\ \\
\textbf{Radiology reports} are written by the radiologist to communicate findings and conclusions from diagnostic imaging to the physicians (specialists or GPs) that requested them.\\ \\
\textbf{Discharge letters } are intended to inform GPs and other physicians on a hospital admission episode for a given patient. Both the Radiology reports and Discharge letters originate from the Erasmus Medical Center only, and are grammaticaly well-formed and easier to understand by others (especially the Discharge letters) compared to the GP entries and the specialist letters.\\ \\
The following are representative examples of each record type in the Erasmus Medical Center Dutch Clinical Corpus \parencite{Afzal2014}. An approximate English translation is printed underneath; the annotated medical terms are in \textbf{bold}.

\subsubsection{GP Entries}

\fbox{\begin{minipage}{\textwidth}
al 6 d temp 39- 40 graden, \textbf{hoesten}, neusverkouden, eet wel OMA re , li een buisje, rustig, forse rhonchi diffuus mogelijk viraal, maar gezien de lange duur nu toch ab
\end{minipage}}

already 6 d (\textit{days}) temperature 39-40 degrees, \textbf{coughing}, runny nose, does eat Otitis Media Acuta ri (\textit{right}), le (\textit{left}) a tube, calm, considerable rhonchi diffuse possibly viral, but considering the long duration now after all ab (\textit{antibiotics})

\subsubsection{Specialist letters}

\fbox{\begin{minipage}{\textwidth}
PED Sinds 3 jr is \#Name\# regelmatig \textbf{misselijk} en braakt hij frequent, gerelateerd aan \textbf{stress} en spanningen; Volledig bloedbeeld en bezinking: g.a.; HP negatief.
\newline \textbf{Coeliakie} neg.
\end{minipage}}

PED (\textit{pediatrics}) Starting 3 yr (\textit{years}) ago \#Name\# (redacted) is regularly \textbf{nauseous} and vomits frequently, related to \textbf{stress} and tension; Complete blood count and sedimentation; n.a. (no abnormalities); HP (\textit{Helicobacter Pylori}) negative. \textbf{Celiac disease} neg. (\textit{negative})

\subsubsection{Radiology reports}

\fbox{\begin{minipage}{\textwidth}
KLINISCHE INFORMATIE: Op 3 januari plotseling \textbf{hoofdpijn}.
\newline Toch \textbf{intracraniale bloeding}?
\newline VRAAGSTELLING: Toch \textbf{aneurysma}?
\newline Er is multifocaal \textbf{spasme} in linker A. cerebri posterior en rechter A. cerebelli superior.
pericallosa distaal is multifocaal \textbf{spasme}.
\newline Differentiaal \textbf{diagnose} moet gesteld worden tussen  \textbf{vasculitis} en  \textbf{spasme} door subarachnoidale  \textbf{bloeding}.
\newline Een  \textbf{aneurysma} is niet aangetoond.
\end{minipage}}

\textit{CLINICAL INFORMATION: On January 3rd suddenly  \textbf{headache}.
\textbf{Intracranial hemorrhage} after all?
INDICATION: \textbf{Aneurysm} after all?
There is multifocal \textbf{spasm} in left A. cerebri posterior en right A. cerebelli superior.
pericallosa distal is multifocal \textbf{spasm}.
Differential \textbf{diagnosis} should be made to distinguish  \textbf{vasculitis} from \textbf{spasm} because of subarachnoidal  \textbf{hemorrhage}.
There is no evidence for an  \textbf{aneurysm}.}

\subsubsection{Discharge letters}

\fbox{\begin{minipage}{\textwidth}
De \textbf{diagnose} distorsie mediale band werd gesteld en patiente kreeg een gipskoker voor 4 weken, daarna fysiotherapie.
\newline Daarna trad \textbf{zwelling} op.
\newline De linker knie toont enige kapselzwelling maar geen \textbf{hydrops}.
\newline \textbf{Diagnose}: Voornamelijk anterolaterale instabiliteit linker knie op basis van een oud voorste kruisbandletsel.
\end{minipage}}

\textit{The \textbf{diagnosis} sprained medial ligament was made and patient got a cast for 4 weeks, followed by physical therapy.
Afterwards \textbf{swelling} occurred.
The left knee is showing some capsule-swelling but no \textbf{hydrops}.
\textbf{Diagnosis}: Predominantly anterolateral instability left knee on the basis of old anterior cruciate ligament injury.}

\newpage
\subsection{RobBERT tables}

We see no consistent differences between the different RobBERT versions in this benchmark (Table \ref{tab:robbert_comparison}). From Table \ref{tab:robbert_drilldown} we conclude that using an almost ten times smaller maximum sequence length of 32 tokens gives comparable performance for the RobBERT models. Furthermore we see a performance comparable to the rule-based approach if we train only the final dense layer of the domain-specific MedRoBERTa.nl model. 
RobBERT does not benefit from the full 512-token size context window and lowering the context-window to 32 tokens only slightly reduces the accuracy\footnote{when considered over the entities that could be classified} while drastically decreasing the computational requirement. Domain-adapted pre-training leads to a slight improvement.

\begin{table}[ht]
\begin{tabular}{llrrr}
\toprule
             Letter category & Prediction method &  Precision &  Recall &    F1 \\
\midrule
           Discharge letters &           RobBERT &      0.953 &   0.974 & 0.963 \\
           Discharge letters &      RobBERT DAPT &      0.964 &   0.984 & 0.974 \\
           Discharge letters &     MedRoBERTa.nl &      0.973 &   0.966 & 0.970 \\
\midrule
General Practitioner entries &           RobBERT &      0.950 &   0.912 & 0.931 \\
General Practitioner entries &      RobBERT DAPT &      0.941 &   0.923 & 0.932 \\
General Practitioner entries &     MedRoBERTa.nl &      0.943 &   0.923 & 0.933 \\
\midrule
           Radiology reports &           RobBERT &      0.960 &   0.963 & 0.961 \\
           Radiology reports &      RobBERT DAPT &      0.956 &   0.960 & 0.958 \\
           Radiology reports &     MedRoBERTa.nl &      0.962 &   0.951 & 0.957 \\
\midrule
          Specialist letters &           RobBERT &      0.934 &   0.890 & 0.911 \\
          Specialist letters &      RobBERT DAPT &      0.945 &   0.905 & 0.924 \\
          Specialist letters &     MedRoBERTa.nl &      0.936 &   0.915 & 0.925 \\
\bottomrule
\end{tabular}
\caption{Comparison of different RobBERT/RoBERTa versions, using a batchsize of $32$, $3$ epochs, a gradient of $10^{-4}$ and a maximum sequence length of $512$.}
\label{tab:robbert_comparison}
\end{table}
\begin{table}[ht]
\begin{tabular}{lrrr}
\toprule
Prediction method &  Precision &  Recall &    F1 \\
\midrule
          RobBERT, 3 epochs, batch size 64, 512 tokens &      0.94 &   0.90 & 0.92 \\
          RobBERT, batch size 32, 512 tokens &      0.95 &   0.94 & 0.95 \\
    RobBERT DAPT, batch size 64, 512 tokens  &      0.94 &  0.93 & 0.94 \\
    RobBERT DAPT, batch size 32, 512 tokens  &      0.95 &  0.95 & 0.95 \\
    MedRoBERTa.nl,  batch size 64, 512 tokens &      0.95 &   0.92 & 0.94 \\      
    MedRoBERTa.nl,  batch size 32, 512 tokens &      0.96 &   0.94 & 0.95 \\
\midrule
    RobBERT, 3 epochs, batch size 32, 32 tokens &      0.95 &   0.93 & 0.94 \\
    RobBERT DAPT, batch size 32, 32 tokens &    0.95 &   0.95 & 0.95 \\
    MedRoBERTa.nl, batch size 32, 32 tokens &     0.96 &   0.94 & 0.95 \\
\midrule
    RobBERT simple-fit, 10 epochs, batch size 128, 512 tokens &      0.89 &   0.65 & 0.75 \\
    RobBERT DAPT simple-fit, batch size 128, 512 tokens &       0.90 &   0.75 & 0.82 \\  
    MedRoBERTa.nl simple-fit, batch size 128, 512 tokens &      0.94 &   0.81 & 0.87 \\ 
\bottomrule
\end{tabular}
\caption{Overview of RobBERT/RoBERTa results}
\label{tab:robbert_drilldown}
\end{table}

\newpage
\newpage
\subsection{DAPT text characteristics}
\label{app:dapt_texts}
\begin{table}[ht!]
\centering
\caption{Text characteristics for domain adapted pre-training}
\begin{tabular}{llll}
\textit{characteristics} & \# of words & \# of unique words & \# of sentences with minimally 3 tokens \\
\hline
Huisarts \& Wetenschap (1957-2019) & 22.839,529 & 510,417 & 1.176,662 \\
NHG directives/standards & 17.052,394 & 257,030 & 921,336   \\
FMS directives & 41.254,869 & 325,357 & 2.188,601 \\
Medical Wikipedia & 4.928,938 & 213,565 & 285,617
\end{tabular}
\end{table}

\newpage
\printbibliography[title={Bibliography}]

\end{document}